# Railway Operation Rescheduling System via Dynamic Simulation and Reinforcement Learning

Shumpei KUBOSAWA[1,2], Takashi ONISHI[1,2], Makoto SAKAHARA[1], Yoshimasa TSURUOKA[1,3]

[1]NEC-AIST AI Cooperative Research Laboratory, National Institute of Advanced Industrial Science and Technology, Tokyo, Japan

[2]Data Science Research Laboratories, NEC Corporation, Kanagawa, Japan

[3]Graduate School of Information Science and Technology, The University of Tokyo, Tokyo, Japan

The number of railway service disruptions has been increasing owing to the intensification of natural disasters. In addition, abrupt changes in social situations such as the COVID-19 pandemic require railway companies to modify the traffic schedule frequently. Therefore, automatic support for optimal scheduling is anticipated. In this study, an automatic railway scheduling system is presented. The system leverages reinforcement learning and a dynamic simulator that can simulate the railway traffic and passenger flow of a whole line. The proposed system enables rapid generation of the traffic schedule of a whole line because the optimization process is conducted in advance as the training. The system is evaluated using an interruption scenario, and the results demonstrate that the system can generate optimized schedules of the whole line in a few minutes.

## 1. Introduction

Since the 2010s, owing to severe natural disasters such as earthquakes and floods, and sudden changes in social situations such as the COVID-19 pandemic, the need to reschedule railway traffic, such as planned suspensions, limitation of service areas, and changes in the number of services are rapidly increasing. Natural disasters and changes in social situations are unpredictable, and operating conditions, such as serviceable areas, number of trains, and passenger flow vary enormously. Thus, preparing all the schedules in advance is impractical, and rapid rescheduling is required after the operating conditions are determined. Additionally, various objectives, such as maximizing passenger benefits by reducing travel time and congestion, and minimizing operating costs, should be considered in rescheduling. In today's situations of unexperienced operating conditions and passenger flows, the burden of rescheduling operations is rapidly increasing, and the need for automated assistance is growing.

Studies on the automatic optimization and generation of railway traffic schedules (time-space diagrams of trains) have been conducted for decades. However, most of the existing methods have only been applied to a limited number of actual operations because they are limited to small scale (e.g. number of stations, number of trains), are limited to a part of the route, or require a long time for calculation. However, if the automatically proposed schedule is limited to only a part of the route or period, manual rescheduling for the remaining portions and connecting them are required. For the human commanders, these tasks would be unfamiliar and inexperienced; thus, such systems are considered impractical.

Therefore, there are two technical requirements for railway traffic rescheduling. One is to schedule the optimal traffic of the entire route and the other is to quickly generate schedules at the time of use. In recent years, artificial intelligence (AI) technologies, such as machine learning, have made remarkable progress in optimization technology due to improvements in computational performance. In this study, we propose a system that uses AI technology and a simulator (Fig. 1) to optimize the railway traffic rescheduling for the entire route and quickly generate a schedule during operation. In the evaluation experiment, we simulated a virtual route (hereafter referred to as "Route X") that describes a double-track line in the Tokyo metropolitan area, as shown in Fig. 1, and demonstrated that quick generation of railway traffic schedule of the whole route that addresses the abrupt occurrence of impassable situations.

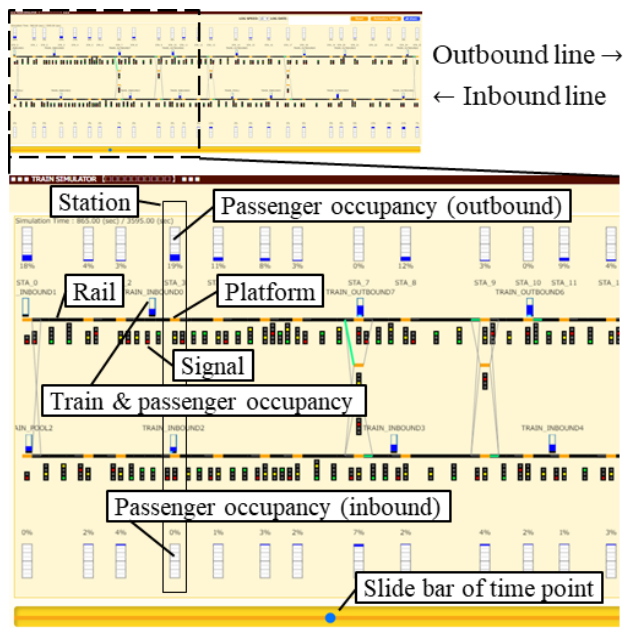

Fig. 1  Visualized whole-route simulation (route X)

## 2. Related work
### 2.1 Railway Schedule Optimization using MILP

Mixed Integer Linear Programming (MILP), one of the most common methods in mathematical optimization, is often used in existing research[1] to automatically generate railway traffic schedules that are optimized for passenger benefits in the event of schedule disruption. In MILP-based schedule optimization methods, for instance, the optimization problem is formulated by representing the arrival and departure times of each train at each station as variables to be obtained, the arrival interval as the linear objective function to be optimized (i.e. to achieve the desired interval), and the route structure, such as the order of each train and station, and the operating conditions, such as the minimum stopping time, as inequalities. The many variables, inequalities created in this way are input to the optimization software (solver) as well as the objective function and calculate to obtain the optimal solution of the arrival and departure times of each train. Using the solution, a time-space diagram of trains can be visualized.

The advantage of MILP-based schedule optimization is that the optimality of the generated schedule is guaranteed by the performance of the solver. On the other hand, the number of variables and inequalities increases as the route size and planning period is expanded. This exponentially increases the computation time of the solver and thus the method is incapable of real scale problems. Even if we employ a quantum annealing computer, which is expected to perform high-speed computation, the MILP problem requires much more qubits (i.e. requires a larger scale of quantum computers) than is currently developed. Additionally, we need to discover a method to reduce the MILP problem of traffic schedule optimization to a Quadratic Unconstrained Binary Optimization (QUBO) problem that can be solved constantly by quantum annealing computers. Thus, the MILP-based method still has some issues for practical use.

## 2.2 Dynamic Simulation

Dynamic simulation is a method to predict the behavior of a dynamical (time-varying) system and is commonly employed in the field of control engineering. To predict time-varying states of a system, a dynamic simulation uses the rules of dynamical changes in states (dynamics) commonly described as differential equations and numerical integration on a time axis starting from an appropriate given initial state. In the field of railway operation, a method using dynamic simulation that reproduces train operation and the flow of each passenger to estimate the benefits of individual passengers and evaluate railway traffic schedules regarding the benefits has been proposed[2].

## 2.3 Reinforcement Learning

Reinforcement learning (RL) is a branch of machine learning and has been studied since the 1970s. RL has received renewed attention after the RL-based Go (an oriental board game) player "AlphaGo" defeated a human professional player in 2015. RL is a method to automatically construct optimal control rules (controller or policy function) for dynamical systems from a given objective function (reward function) and input/output data of the system. The optimal control rules refer to "a function that takes the current observed state as input and outputs an action to achieve the given desired state." In RL, the target system is automatically manipulated to collect input and output data, the optimality of each output is evaluated by a reward function, and the collected data is used to learn control rules that improve optimality in the future. Even in a system such as Go games where the evaluation is not determined until the end of the game, RL is possible to construct the optimal method to play (the optimal control rules) because the optimization can reflect far future state changes.

### 2.3.1 Domain Randomization

RL is a data (experience) driven machine learning method. High optimality can be expected in the proximity of situations experienced during learning (interpolation), but optimality in unexperienced situations (extrapolation) is generally not guaranteed. Thus, to address a wide range of situations, these situations should be experienced during learning. For this purpose, domain randomization has been proposed, in which the simulation parameters (e.g. suspended area, suspended interval, and simulation start time) are changed randomly in each episode and vary situations during training.

## 2.4 Behavioral Cloning

Behavioral cloning (BC) is another automatic method to construct control rules for a system. In BC, control rules are extracted from recorded data of human operations. BC is a supervised learning method that approximates a policy function (control rules) with the observable state at each time as the input and the performed manipulation as the output. BC is used to reproduce the current control rules when optimization by search of manipulation is impractical owing to complex dynamics of the system i.e. difficulty in predicting state changes. In addition to BC, inverse reinforcement learning (IRL) is also known as a method to reproduce the existing control rules. In IRL, the reward function (i.e. objective of the operation) is estimated from the operation record and reinforcement learning (value and policy estimation) is performed for the estimation. However, IRL is computationally expensive because it involves reinforcement learning repeatedly. Besides, the resulting reward functions are often difficult to interpret. On the other hand, BC is a realistic method for constructing control rules for complex systems because it is a simple supervised learning method; thus, the computational complexity is relatively low.

However, since BC extracts control rules from actually experienced data, optimality of unexperienced (unrecorded) situations is difficult to improve (extrapolation). Additionally, since BC is also a machine learning method, a large amount of quantitative (digital) data is required. On the other hand, to improve the optimality of a given objective in RL, the system is automatically manipulated to collect the required data in large quantities. Therefore, the optimality of RL can be improved more than BC.

## 3. Proposed Method

To quickly generate optimized railway traffic schedules for the whole route during operation, we propose a system that combines a dynamic simulator capable of reproducing train and passenger movements along a whole route of a railway line and RL. The system can address various objective functions for optimization via RL and various situations via domain randomization. To improve the learning speed and to reproduce realistic situations, we developed a railway traffic simulator that is suitable for reinforcement learning.

## 3.1 RL-Oriented Railway Simulator

The developed simulator is capable of reproducing train movements complied with general railways signaling systems, such as signaling block and interlocking, and also passenger flow. Fig. 1 shows a screenshot. The detailed specification of the railway line including the distance of each block, detailed path inside stations, limit speed can be configured on the simulator. The configuration is used in calculating dynamically changing speed limit (signal indication) at each location. Thus, the simulator enabled simulations with an accuracy of seconds. Additionally, passenger flow between trains and inside the station, and between inside and outside the station can be simulated. The symbol of the train in Fig. 1 is a bar graph, showing the ratio of passengers to capacity. By simulating the passenger flow, the objective of scheduling optimization can involve the number of passengers staying in the station and the degree of train congestion. The manipulation points of the simulator are clear/go (on) or stop (off) for the departure and junction signals. Trains are operated automatically according to signal indications and in compliance with speed limits. This enabled to reproduce a wide range of possible situations according to the route facilities. Additionally, a function to cause impassable situations for any period and location is implemented to reproduce the accidental impassable situations.

To enable the automatic operation of this simulator by RL, we implemented an API that conforms to the OpenAI Gym interface commonly used in RL. To collect a large amount of data required for RL through a large number of simulations, the simulator is designed to run in parallel to improve collection efficiency. The route model that represents the facilities of each railway line is described in the configuration file for the route. Thus, the simulator supports arbitrary facilities such as single-track, double-track, and quadruple-track railway lines and their combination by replacing the configuration file.

## 3.2 Rule-based Hierarchical Learning Environment

In this simulator, all junction signals and departure signals are operable. In the case of route X, there are 88 manipulation points. In RL in general, as the number of manipulation points increases, the number of action candidates (the number of combinations) increases exponentially and the learning (search) efficiency rapidly deteriorates. This problem is commonly regarded as the "curse of dimensionality." Regarding the presented simulator, there are a large number of superfluous manipulation candidates that are possible but do not affect the train movement (e.g., turning on the departure signal when the train is not on the platform). Additionally, from the perspective of scheduling traffic of the entire route, macro decisions such as "whether to turn back or go straight" at a certain station are important, but micro-decisions such as "which platform to enter" may not be important because there is little room for optimization. In this study, we adopted a hierarchical environment in which the optimization is limited to macroscopic decision-making policy (control rules), while microscopic decision-making is automated with given rules and not to learn. This structure is shown in Fig. 2. The RL agent (learning controller) makes macroscopic decisions, such as "whether to turn back or go straight ahead" and stopping time, while microscopic decisions, such as detailed turning procedures and managing train departures, are made by the RB agent (automatic controller), which is described in a rule base (RB). This method reduced the number of manipulation points to be trained to seven in the route X case.

## 3.3 Distributed and Cooperative RL

After disruption of the train schedule, an optimal traffic rescheduling for the operable section is required if the impassable section remains. These impassable situations owing to accidents are often resolved in a few hours. After the accidental situation is resolved, to smoothly provide train information to passengers and to assign train crews and cars to each train, recovering the original timetable schedule is necessary. For this reason, the objectives of railway traffic management involving accidental impassable situations can be distinguished in chronological order as follows:

1. In normal times, the objective is to operate on time.
2. When an impassable situation occurs, the objective is to minimize passenger inconvenience (disruption response).
3. After the impassable situation is resolved, the objective is to recover the planned timetable.
4. After the planned timetable is recovered, the objective is to operate the trains on time (return to 1.).

By extracting the available paths at each time (operation rules) from the given timetable and the path structure in each station, the control rules that operate the simulator according to the planned timetable (1., 4.) can be constructed as a rule-based timetable agent that manipulates each signal on time. The remaining two control rules should be constructed by optimization of each objective: (2.) the control rules for dealing with impassable situations and (3.) the control rules for recovering the normal schedule.

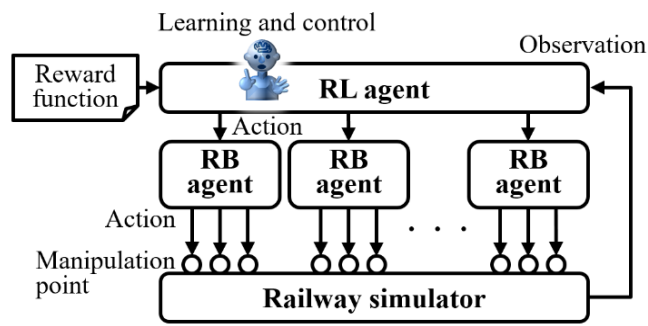

Fig. 2 Schematic of hierarchical learning architecture

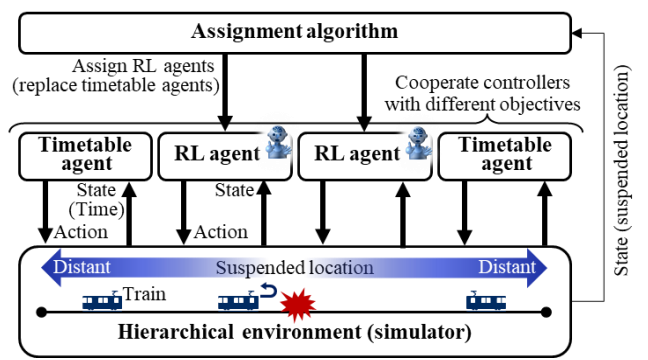

Fig. 3 Schematic of distributed cooperative control

If the RL agent is solely required to deal with impassable situations (2.), the RL agent can minimize the congestion and the stopping time of trains by freely manipulating the simulator. However, the smooth switchover from one controller (e.g. RL agent for addressing impassable situations (2.)) to another controller (e.g. the timetable agent) is a nontrivial problem. Particularly, starting from arbitrary distributions of trains on the route, the task of restoring the time-varying positions of a large number of trains to a specified situation (i.e., recovering a timetable) in a short time is highly complex and difficult problem. Even human expert commanders would not deal with such a complicated problem with a highly large freedom of degree.

In this study, we propose a method of learning control rules that can both cope with impassable situations and restore the timetable by utilizing the characteristic of impassable accidents that "disruptions occur locally around the accidental point and delays spread from there." At the beginning of the impassable situation, disruptions occur in the vicinity of the accidental section but do not occur in the distant sections yet. In the vicinity of the accidental section, restoring the planned schedule is impossible until the situation is resolved, so rescheduling traffic to minimize train stoppage time and congestion is required. However, in distant areas, continual operation according to the original schedule would be possible. Additionally, the delay will not propagate and the distant schedule can be maintained if appropriately decided to turn back the train or bring the train to the depot when a train arrives in the vicinity of impassable sections from a distant place. Utilizing this characteristic, our method uses an RL-based rescheduling agent in the vicinity of the disrupted area, while the timetable agent continues to operate in the distant area. This configuration is shown in Fig. 3. To determine the vicinity, i.e. the assignment of control rules, a rule-based assigner is constructed in advance, based on the layout of turnaround stations and train depots on the route, and the assigner determines which manipulation points should be replaced to an RL agent from the timetable agents.

### 3.4 Training Configuration

We used Proximal Policy Optimization (PPO) as the learning algorithm for reinforcement learning of the rescheduling agent. The reward function for responding to impassable situations while considering the recovery of the planned schedule is a weighted linear sum of optimization indices such as maximization of train travel speed, minimization of congestion, and minimization of the difference between the distribution of the current trains and that of the planned schedule. The rescheduling agent can observe the existence and location of impassable sections. This enables the agent to determine whether there is an accident and to learn the operation to recover the original schedule after the impassable situation is resolved. Additionally, we introduced domain randomization to randomly change the location, time, and duration of the impassable situation during training. This allows a single system to address impassable situations at any location and at any time of the day.

### 4. Experimental Evaluation

In this paper, we describe the evaluation results of the method proposed in §3.2, which uses reinforcement learning and hierarchical training environments. To evaluate the schedules proposed by the method when impassable situations occur, we conducted simulation experiments on route X shown in Fig. 1. Route X is a totally double-track line with 26 stations and 4 intermediate stations that can be turned around. There are 16 trains on the route. We configured the appropriate origin-destination matrix of incoming passengers from outside of each station and simulated the passenger flow.

### 4.1 Rescheduling via RL and Hierarchical Environment

The reward function for learning the rescheduling agent was set to maximize the number of passengers arriving at their destination and minimize the stoppage time. For comparison, we also simulated the case without rescheduling, in which the trains are operated according to the original schedule even after an impassable situation occurred. Fig. 4 and Fig. 5 show the time-space diagrams of trains for the case without and with rescheduling, respectively. In both figures, the vertical axis represents the position and the horizontal axis represents time. The black bold lines indicate the impassable sections and periods. The colorful thin solid lines represent the position of each train. In the area framed by a dotted black line in

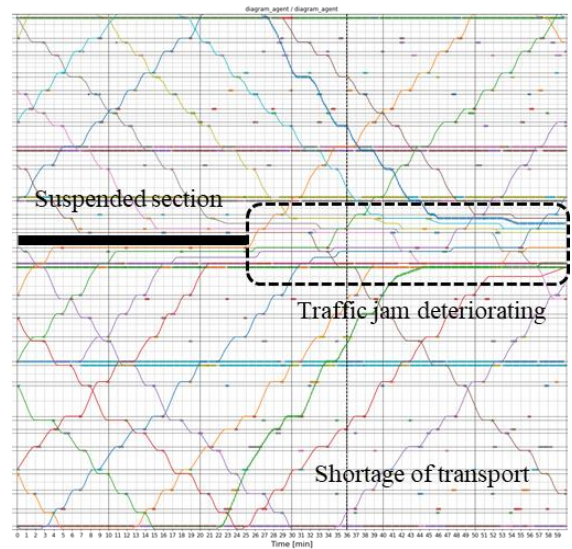

Fig. 4 Trajectories of trains without rescheduling

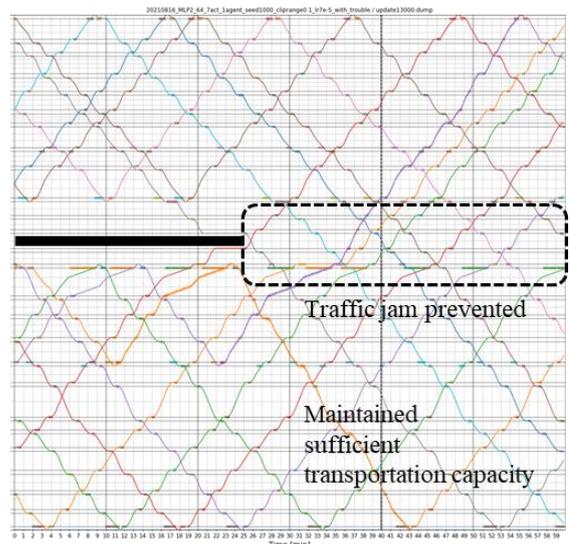

Fig. 5 Trajectories of trains with rescheduling

Fig. 4, many horizontal lines are shown. This indicates that many trains departed for the impassable section and traffic jams i.e. stopping between stations (which significantly increase passenger frustration) occurred around the impassable section during the accidental period. Furthermore, even after resolving the impassable situation, the traffic jam deteriorated because the trains are operated on the original timetable, i.e. these trains depart each station on schedule, without considering the jam. On the other hand, in the dotted-framed area of Fig. 5, there are no long-lasting horizontal lines i.e. traffic jams; thus, Fig. 5 indicates that the rescheduling agent successfully addressed the impassable situation. Additionally, comparing the areas above and below the dotted-framed area in Fig. 4 and 5, the number of trains decreases in the case without rescheduling (Fig. 4), but trains come and go almost regularly in the rescheduled case (Fig. 5). The objective (reward setting) of the rescheduling was to maximize the number of passengers arriving at the destination and minimize the stopping time; thus, this evaluation demonstrated that the proposed method successfully generated a railway traffic schedule that is optimized for the user-specified indices.

### 4.2 Processing Time

In RL methods, the learning (optimization process) is completed before the actual operation; thus, the proposed method can output schedules of a total route using only one episode of simulation (i.e. one series of predictions) during the operation. Thus, the processing time for rescheduling is less than 3 min for the scale of route X, and even less than 10 minutes for the entire sections of a complex quadruple line. Notably, in conventional scheduling methods utilizing mathematical programming such as MILP, this overwhelmingly fast processing for those large-scaled problems would be impossible i.e. the calculation never finishes in the practical period. The processing time of the proposed method is expected to be reduced further in the future by improving the simulator.

### 5. Conclusion

In this study, we proposed a system that can quickly generate a railway traffic schedule of a total line in impassable situations using a dynamic simulator and reinforcement learning. We also developed an RL-oriented simulator for railway traffic and passenger flows. In the evaluation experiment, we showed that reinforcement learning using the hierarchical environment can optimize schedules of the entire route for a specified objective function. We are proceeding to evaluate the performance of recovering the original schedule after resolving impassable situations using distributed cooperative reinforcement learning.

Command operations of railway traffic management require quick response to various situations such as planned suspensions, limitation of service areas, and changes in the number of services owing to sudden and unforeseeable events. These operations are conducted by experienced commanders with expertise and professional skills because there are many factors to consider and a variety of situations are possible. Owing to the large complexity of these tasks, automatic support has been impractical conventionally. In recent years, however, improvements in computational performance have made it possible to reproduce and evaluate the traffic situation at high speed. Additionally, AI technologies such as optimization by machine learning are achieving a practical level. With this background, we have been developing a railway traffic rescheduling system based on machine learning techniques. As the next step of this research, we intend to contribute to the operation optimization of railway companies, which are responsible for social infrastructure, by promoting the implementation of these technologies in actual operation.

# ダイナミックシミュレーションと強化学習による
# 運転整理ダイヤ生成システム

○窪澤　駿平　（産総研・ＮＥＣ）　　大西　貴士　（産総研・ＮＥＣ）

坂原　誠　　（産総研）　　　　　　鶴岡　慶雅　（産総研・東大）

## Railway Operation Rescheduling System via Dynamic Simulation and Reinforcement Learning

Shumpei KUBOSAWA[†], Takashi ONISHI[†], Makoto SAKAHARA, and Yoshimasa TSURUOKA[‡]

NEC-AIST AI Cooperative Research Laboratory, AIST, Aomi 2-4-7, Koto-ku, Tokyo

[†]NEC Corporation, [‡]The University of Tokyo

The number of railway service disruptions has been increasing owing to intensification of natural disasters. In addition, abrupt changes in social situations such as the COVID-19 pandemic require railway companies to modify the traffic schedule frequently. Therefore, automatic support for optimal scheduling is anticipated. In this study, an automatic railway scheduling system is presented. The system leverages reinforcement learning and a dynamic simulator that can simulate the railway traffic and passenger flow of a whole line. The proposed system enables rapid generation of the traffic schedule of a whole line because the optimization process is conducted in advance as the training. The system is evaluated using an interruption scenario, and the results demonstrate that the system can generate optimized schedules of the whole line in a few minutes.

*Keywords:* railway operation rescheduling, optimization, automated planning, reinforcement learning, dynamic simulation

## １．はじめに

2010 年代以降, 地震や水害などの自然災害の激甚化や, COVID-19 パンデミックなどにみられる社会情勢の急変により, 計画運休や運行エリアの縮小, 運行本数の変更など, 運行計画の再立案業務が増加している。自然災害や社会情勢の変化は予測困難であり, また運行可能区間や本数, 旅客流動等の運行条件は多岐に亘る。このため, 運行計画を事前に全て列挙して準備するのは現実的でなく, 運行条件の決定後に迅速に立案する必要がある。加えて, 計画立案にあたっては, 移動時間短縮や混雑低減などの旅客便益最大化や, 運行コスト最小化など, 多様な観点での最適化が求められる。未経験の運転条件や旅客流動が発生している現代では, 計画作業の負荷が急増しており, 自動化による支援の必要性が高まっている。

運行計画（ダイヤグラム）を自動で最適化・作成する研究は長年進められてきた。しかし, 既存手法の多くは, 対応できる路線規模（例：駅数, 列車数）が小規模であり, 路線の一部区間の運転整理案に限られる, あるいは計算に長時間要するなどの理由から, 実運用への導入事例は限定的である。司令（指令）員は, 路線全体の状況をイメージして運行計画を立てるが, 自動提案されるダイヤ案が一部区間に限られると, その区間を除いた区間だけに限定した計画立案や, 前後の区間を接続するための計画立案という, 従来の司令業務に無い新たな作業が必要になるため, 実際の業務において使いづらく感じられてしまう等の課題が考えられる。

このため, ダイヤグラム最適化における技術的要件は, 路線全区間の運転計画立案ができることと, 利用時に迅速に立案できることの２点である。最適化技術に関しては, 近年では計算性能の向上により, 機械学習など人工知能（AI）技術の発展が著しい。そこで本研究では, AI 技術とシミュレータ（Fig.1）を利用して, 路線全区間の運行計画を最適化し, 運用時には迅速にダイヤグラムを生成するシステムを提案する。評価実験では, Fig.1 に示した首都圏の複線１路線の全区間を想定した仮想路線（以降, 路線 X と呼ぶ）のシミュレーションを行い, 不通区間発生時の運転整理ダイヤの最適化と, 路線状況に応じた迅速なダイヤグラム生成が可能であると示した。

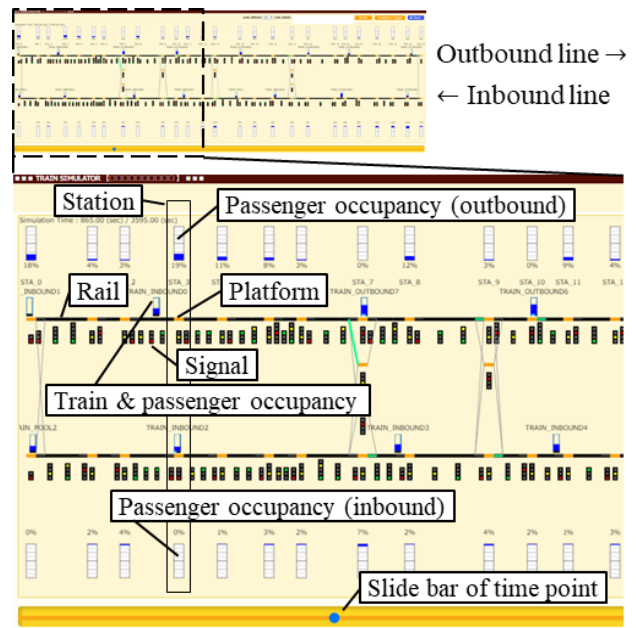

Fig.1　Visualized whole-route simulation (route X)

## 2．関連研究
### 2.1 混合整数計画法によるダイヤ最適化
ダイヤ乱れ発生時の運転整理ダイヤや，旅客の便益について最適化した計画ダイヤを自動生成する既存研究[1]では，最適化における一般的な手法の一つである混合整数計画法（Mixed Integer Linear Programming; MILP）が用いられることが多い。MILPによるダイヤ最適化では，解きたい最適化問題として，各列車の各駅の着発時刻などを求めたい変数として表し，列車到着間隔などを最適化したい目的関数として表し，さらに各列車や各駅の順序などの路線構造や，最低停車時間などの運行条件を不等式で表現して立式する。こうして作成した多数の変数や不等式を，最適化ソフトウェア（ソルバー）に入力して計算し，各列車の着発時刻を最適解として得る。得られた着発時刻の図示によりダイヤグラムが得られる。

MILPによるダイヤ最適化では，生成されたダイヤ（解）の最適性が，ソルバーの性能により保証されるメリットがある。一方で，路線規模や計画期間を拡大すると，変数や不等式の個数が増える。これによりソルバーの計算時間が指数関数的に増加し，現実規模の問題では計算が終わらなくなる課題がある。高速計算が期待される量子コンピュータ，特に実用化の進むアニーリング方式を用いるにも，同様の問題では，現在実用化されているよりも大規模な量子コンピュータ（多数の量子ビット）が必要である。加えて，ダイヤ最適化のMILP問題を，量子アニーリングで解ける無制約二値二次最適化（Quadratic Unconstrained Binary Optimization; QUBO）問題かつ，安定的に解ける問題に還元する方法が新たに必要である。この様に，MILPによる方法には実用化への課題が残る。

### 2.2 ダイナミックシミュレーション
システムの時間変化を再現・予測する方法に，ダイナミックシミュレーションがある。ダイナミックシミュレーションは，状態の時間変化の法則（ダイナミクス）を微分方程式などで記述し，適当な所与の初期状態を起点として，時間軸の数値積分などにより時間変化する状態を再現・予測する手法である。鉄道運行分野では，列車運行と旅客一人一人の移動を再現したダイナミックシミュレーションにより，個々の旅客の便益を推計し，便益に関してダイヤを評価する手法[2]が提案されている。

### 2.3 強化学習
機械学習の一分野に強化学習（RL）がある。1970年代から研究されてきた手法だが，2015年に囲碁のプロ棋士を破ったAlphaGoというAIプレーヤの開発で利用されたことで再び注目を集めている。RLは，ダイナミカルシステムの最適制御則（コントローラ・方策関数）を，所与の目的関数（報酬関数）と，システム（例：シミュレータ）の入出力データから自動構築する手法である。最適制御則とは，「現在の状態を入力として，所望の状態に到達するための操作を出力する関数」を指す。RLでは，システムを自動操作して入出力データを自動収集し，報酬関数で出力の最適性を評価し，収集したデータを用いて「将来に亘る最適性が向上する操作方法」を機械学習する。将来の変化を考慮して最適化するため，囲碁の様に劣勢や優勢を繰り返しつつ，対局が終わるまで勝敗（評価）が定まらないシステムであっても，最適な打ち方（操作方法），すなわち最適制御則を構築できる。

#### 2.3.1 Domain Randomization
RLは，データ（経験）に基づく機械学習を利用する手法である。このため，学習時に経験した状況の近傍では高い最適性（内挿性）が期待できるが，未経験の状況における最適性（外挿性）は一般に保証されない。このため，幅広い状況に対応させるには，幅広い状況を学習時に経験させる必要がある。このための方法として，学習時にシミュレーションのパラメタ（例：不通区間，シミュレーション開始時刻）を毎回ランダムに変化させるdomain randomizationという手法が提案されている。

### 2.4 Behavioral Cloning
RLの他にも，システムの制御則を構築する方法のひとつに，人間の操作記録から制御則を抽出するbehavioral cloning（BC）という手法がある。BCは，各時刻の状態を入力，実施された操作を出力とした制御則を，操作記録データから関数近似する教師有り学習である。BCは，システムのダイナミクスが複雑で状態変化の予測が困難である等により，操作の探索による最適化が困難な場合に，現状の制御則を再現する目的で利用される。BCの他にも，制御則を再現する方法として，操作記録から報酬関数を推定し，同時に強化学習（方策推定）を行う逆強化学習も知られている。ただし，逆強化学習は強化学習を内包するために計算量が多く，また得られる報酬関数も解釈が困難である場合が多い。一方で，BCは教師有り学習であり計算量が比較的に少ないため，複雑なシステムの制御則構築においては現実的な方法である。

ただし，BCは既に存在するデータ（例：実績データ）から制御則を抽出するため，実績を上回る最適性の向上や，未経験の状況における最適性（外挿性）は期待できない。また，BCも機械学習である以上，大量の数値化された実績データが必要である。一方RLでは，必要なデータを，システムを自動操作して新たに大量に収集しつつ，所定の指標に関して最適性を高めてゆく。このため，BCよりもRLの方が最適性を向上することができる。

## 3．提案手法
本研究では，鉄道路線の全区間の列車と旅客の移動を再現可能なダイナミックシミュレータと，RLを組み合わせることで，路線の全区間に対応し，さらに様々な状況や目的関数について最適化された運行計画案を，運用時には迅速に作成するシステムを提案する。本研究では，RLに適したシミュレータを新たに開発して使用した。

### 3.1 強化学習用シミュレータ
開発したシミュレータは，列車の移動をはじめとして，閉塞や連動（インターロック）など，一般的な鉄道信号設備の再現に加えて，旅客流動も再現可能である。Fig.1にスクリーンショットを示す。各閉塞の距離や駅構内の進路を精密に再現することで，信号によりダイナミックに変化する制限速度を列車移動に反映し，秒単位の精度によるシミュレーションを可能とした。また，列車と駅場内，駅場内と駅場外の旅客流動も再現可能である。Fig.1の列車のマークは棒グラフになっており，定員に占める乗客の比率が表示されている。旅客流動を再現することで，運行計画最適化における目的関数に，駅場内の滞留人数や，列車の混雑度など，旅客流動に関する観点も設定可能とした。シミュレータの操作箇所は，出発信号と場内信号のOn（進行）とOff（停止）である。列車は，信号現示に従って，制限速度を遵守して自動で運転される。これにより，路線設備として実施可能な状況を幅広く再現可能とした。加えて，不通再現のため，任

意の閉塞を任意の期間で通行不能とする機能も実装した。
　本シミュレータは，RL による自動操作を可能とするため，RL で一般的な OpenAI Gym インタフェースに準拠した API を用意した。また，RL に必要な大量のデータを膨大な回数のシミュレーションにより収集するため，収集効率向上のために並列動作するよう設計した。各路線の設備を表す路線モデルは，設定ファイルに記述して使用する。これにより，複線や複々線など任意の構造の路線に，設定ファイルの切り替えで対応可能とした。

### 3.2 ルールベース階層型強化学習環境

　本シミュレータでは，全ての場内信号と出発信号が操作可能である。このため，例えば路線 X の規模で 88 個もの操作箇所がある。RL では一般に，「次元の呪い」として，操作箇所数が増えると操作の候補数（組み合わせ数）が指数関数的に増加し，学習（探索）効率が急速に低下することが知られている。一方で，可能な操作だが，列車移動に影響しない無用な操作（例：列車がホームに無いのに出発信号を On にする）も大量に存在する。また，路線全体の運行計画立案という観点からは，ある駅で「折り返すか直進させるか」などのマクロな意思決定は重要だが，「どちらのホームに進入させるか」などのミクロな意思決定は，最適化の余地が小さく重要でない場合がある。そこで本研究では，最適化するのはマクロな意思決定方策（制御則）に絞り，ミクロな意思決定は所与のルールで自動化して学習対象外とする階層型環境を採用した。この構造を Fig.2 に示す。RL エージェント（学習制御器）は，「折り返すか直進させるか」や停車時間など，マクロな意思決定を行い，詳細な折り返し手順の実施や列車の出発管理などのミクロな意思決定は，ルールベース（RB）に記述して自動化した RB エージェント（自動制御器）が担う。これにより，路線 X の場合には学習対象の操作箇所数を 7 個にまで削減した。

### 3.3 分散協調型強化学習

　事故等による不通発生後の運転整理においては，不通区間が存在する期間は，運転可能区間の最適運行計画を立案する必要がある。一方で，事故等による不通は，数時間程度で解消されることが多い。不通解消後は，旅客への列車案内や，乗務員・車両手配などを円滑に行うため，計画ダイヤ，すなわち時刻表通りの運転を迅速に回復する必要がある。このため，運行制御の目的は，時系列順に次の様に区別できる：

1. 平常時には，時刻表通りの運行が目的，
2. 不通発生後は，旅客不便最小化（不通対応）が目的，
3. 不通解消後は，計画ダイヤ回復が目的，
4. 計画ダイヤ回復後には，時刻表通りの運行が目的。

　計画ダイヤ通りにシミュレータを操作する制御則(1., 4.)は，所与の時刻表に記載された各駅への着発時刻と各駅の進路構造から，時刻毎に開通する進路（運行ルール）を抽出することで，時刻を入力として進路を操作するルールベースの時刻表エージェントとして構築できる。このため，運転整理に必要な制御則は，(2.)不通対応目的の制御則と(3.)ダイヤ回復目的の制御則の 2 個である。

　不通対応（2.）だけならば，RL エージェントにシミュレータを自由に操作させることで，停車時間最小化や混雑最小化などの適当な指標に関して学習は可能である。一方で，別の制御則（時刻表エージェント）への円滑な切り換えが必要な，計画ダイヤ回復(3.)を最適化する方法は自明ではない。特に，時間変化する多数の列車の各

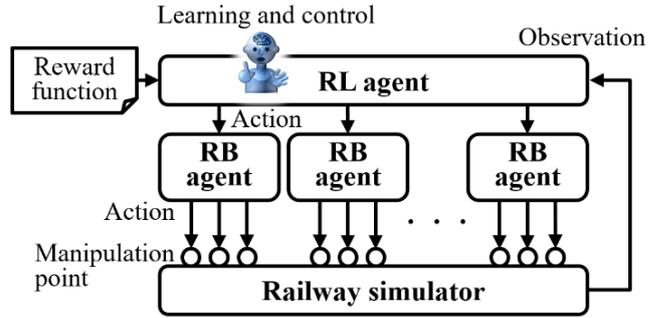

Fig.2　Schematic of hierarchical learning architecture

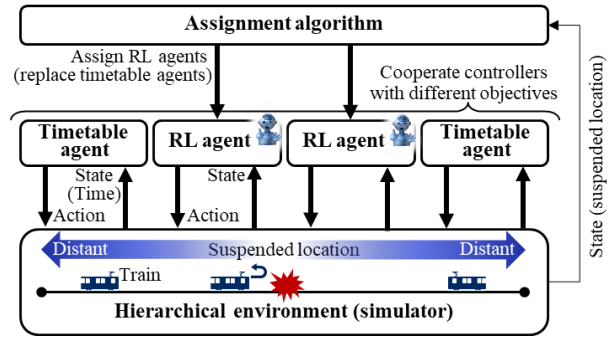

Fig.3 Schematic of distributed cooperative control

位置を，任意の列車配置を起点として，短時間で指定の状況に復元する（ダイヤ回復する）タスクは，複雑性が大きく学習も困難である。人間の司令員も，その様な自由度の大きいタスクは扱っていないと考えられる。

　そこで本研究では，「不通は局所的に発生し，遅延はそこから広がってゆく」という特徴を利用した，不通対応とダイヤ回復を両立する制御則の学習方法を提案する。不通発生当初，不通区間の近傍ではダイヤ乱れが生じるが，遠方の区間ではダイヤ乱れは未発生である。不通区間近傍では，不通が解消されるまで計画ダイヤ回復は不可能のため，列車停車時間最小化などを目的とした運転整理が必要であるが，遠方では引き続き計画ダイヤ通りの運行が可能である。また，不通発生中でも，遠方から近傍に列車が到着した際に，折り返しや入庫など適切に操作できれば，遅延は伝播せず遠方のダイヤは維持可能である。この性質を利用して，本手法では不通発生箇所の近傍では RL により運転整理エージェントを構築して利用し，遠方では引き続き時刻表エージェントに操作させる。この構成を Fig.3 に示す。近傍かどうかの判定，すなわち制御則の割り当てには，折り返し可能駅と車庫保有駅の配置に基づき，不通区間毎にどの操作箇所を運転整理エージェントに置き換えるか判定するルールベース割り当て器を事前に構築して使用する。

### 3.4 学習設定

　運転整理エージェントの強化学習にあたり，学習アルゴリズムには近接方策最適化法（Proximal Policy Optimization; PPO）を用いた。また，計画ダイヤ回復を考慮しつつ不通対応するための報酬関数には，列車移動速度の最大化や混雑最小化など不通対応の最適化指標と，計画ダイヤによる列車位置とシミュレーション上の列車位置との差分最小化など，計画ダイヤを再現するほど高得点となる指標との重み付き線形和を採用した。運転整理エージェントには，不通区間の有無と箇所も入力する。これにより，エージェントは不通の有無を判別可能であ

り，不通解消後には計画ダイヤを回復する操作を学習可能である。加えて，学習時には，不通発生箇所と発生時刻，および継続時間をランダムに変化させる domain randomization を導入した。これにより，1個のシステムで,任意の箇所と任意の時間帯の不通に対応可能である。

## ４．評価実験

本稿では，§3.2にて提案した階層型環境による強化学習を利用した方法の評価結果を述べる。不通区間発生時の運行計画立案について，Fig.1に示した路線Xを題材としてシミュレーション実験を行った。路線Xは全線複線であり，全26駅，中間の折り返し可能駅は4駅存在する。全16編成が在線している。旅客流動も再現し，適当なODにて場外から各駅に入場するよう設定した。

### 4.1 階層型環境と強化学習を用いた運転整理

運転整理エージェントの学習における報酬関数には，目的地に到着した旅客数の最大化と，停車時間の最小化を設定した。比較のため，適当な計画ダイヤを設定し，不通発生後も計画ダイヤ通り運行し続けた運転整理なしの場合の挙動も再現した。Fig.4に，運転整理なしの場合，Fig.5に運転整理ありの場合のダイヤグラムを示す。どちらの図も，縦軸が位置を，横軸が時間を表す。太線の区間および期間は通行不可能である。細い実線が，各列車の位置の時間変化を表す。Fig.4の点線で囲んだ区間・時間帯では，水平の線が多くみられる。これは不通発生中に，不通区間に向けて数多くの列車が出発してしまったため，不通区間の前後で団子状態が発生し，さらに時刻表通りの時間に発着させようとしたために，詰まった列車が適切に到着・出発できずに駅間停車が多数発生している状況を示している。一方で，Fig.5に示す，学習した運転整理エージェントによる制御によると，同じ点線領域の団子状態は存在せず，円滑な運行が示されている。また，点線領域の上下の領域を比較しても，運転整理なしの場合には列車本数が減少しているが，運転整理ありの場合には，ほぼ定期的に列車が往来していることが示されている。運転整理の目標（報酬設定）は，目的地到着旅客数の最大化，および停車時間の最小化であったことから，本手法により，指定した指標に基づく運転整理案の立案が可能であると示された。

### 4.2 処理速度

強化学習では,実運用に供する前に学習（最適化処理）が完了しているため，運用時には1回のシミュレーションのみで路線全体の運転整理案の出力が可能である。出力に要する時間は，路線Xの規模で最大3分程度，複々線規模の路線全区間でも10分未満であり，MILPなどの数理計画による解法では現実的に計算不可能なほど（計算が終了しない）大規模な問題設定にて，圧倒的な高速処理を実現した。処理速度は，シミュレータの改良により，今後もさらなる改善を見込んでいる。

## ５．おわりに

本研究では，鉄道路線全体を再現するダイナミックシミュレータと強化学習の組み合わせにより，運転支障時の運転整理案を迅速に立案するシステムを提案した。シミュレータには，本研究で新たに開発したものを使用した。評価実験では，階層型環境を用いた強化学習により，指定した目的関数について最適化された路線全体の運転

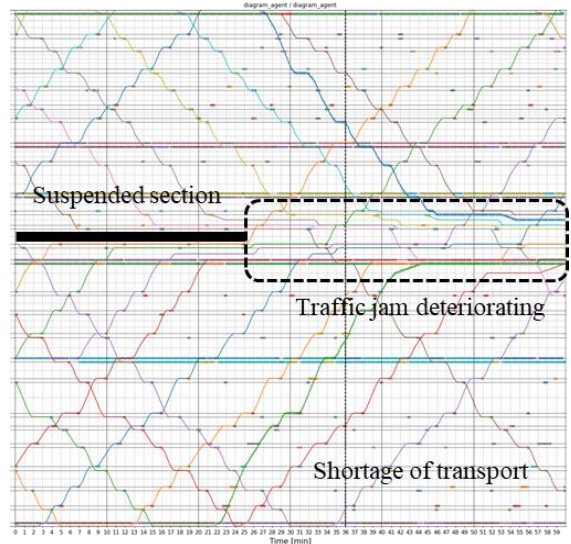

Fig.4 Trajectories of trains without rescheduling

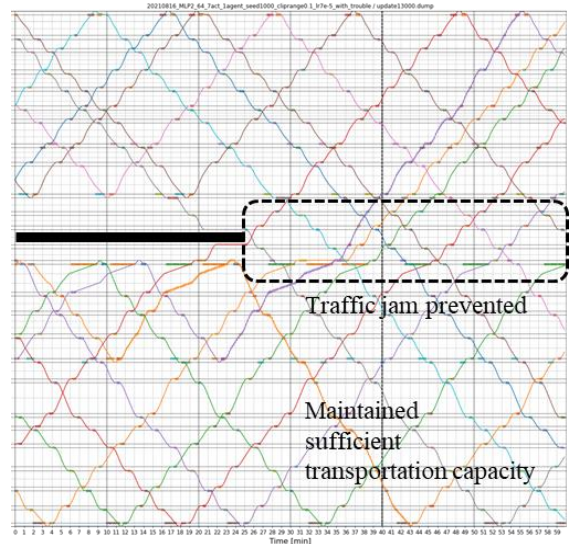

Fig.5 Trajectories of trains with rescheduling

整理案が立案可能と示した。分散協調型強化学習による計画ダイヤ復旧についても，現在検証を進めている。

突発的な事象発生に伴う運転整理や計画運休など，多様な状況への迅速な対応が求められる指令・司令業務は，考慮すべき要素も多いため，豊富な経験と熟練が求められる。また，その複雑性の大きさから，従来は自動支援が困難であった。しかし近年では，計算性能の向上により，状況の再現・評価を高速に行える様になってきた。また，機械学習による最適化などの人工知能技術が実用レベルに到達しつつある。こうした背景から，本研究では機械学習を応用したダイヤグラム最適化技術の開発を進めてきた。今後は，本技術の実運用における検証を推進することで，社会インフラを担う鉄道事業者のオペレーション最適化に貢献してゆく所存である。